\documentclass[10pt,twocolumn,letterpaper]{article}

\usepackage{wacv}              

\usepackage{graphicx}
\usepackage{multirow}
\usepackage{amsmath}
\usepackage{amssymb}
\usepackage{booktabs}
\usepackage{algorithm}
\usepackage{algpseudocode}
\algrenewcommand\algorithmicrequire{\textbf{Input:}}
\algrenewcommand\algorithmicensure{\textbf{Output:}}
\newcommand{\Algorithm}[1]{\item[\textbf{Algorithm:}] #1}
\DeclareMathOperator*{\argmax}{arg\,max}

\usepackage[pagebackref,breaklinks,colorlinks]{hyperref}

\usepackage[capitalize]{cleveref}
\crefname{section}{Sec.}{Secs.}
\Crefname{section}{Section}{Sections}
\Crefname{table}{Table}{Tables}
\crefname{table}{Tab.}{Tabs.}
\usepackage[symbol]{footmisc}

\def\wacvPaperID{1194} 
\def\confName{WACV}
\def\confYear{2024}

\begin{document}

\title{Lost Your Style? Navigating with Semantic-Level Approach \\ for Text-to-Outfit Retrieval}

\author{
Junkyu Jang, \quad Eugene Hwang, \quad Sung-Hyuk Park*\\
KAIST, College of Business\\
Seoul, Korea\\
{\tt\small  \{jbkjsm, hegene3686, sunghyuk.park\}@kaist.ac.kr}
}

\maketitle

\begin{abstract}
Fashion stylists have historically bridged the gap between consumers' desires and perfect outfits, which involve intricate combinations of colors, patterns, and materials. Although recent advancements in fashion recommendation systems have made strides in outfit compatibility prediction and complementary item retrieval, these systems rely heavily on pre-selected customer choices. Therefore, we introduce a groundbreaking approach to fashion recommendations: text-to-outfit retrieval task that generates a complete outfit set based solely on textual descriptions given by users. Our model is devised at three semantic levels–item, style, and outfit–where each level progressively aggregates data to form a coherent outfit recommendation based on textual input. Here, we leverage strategies similar to those in the contrastive language-image pretraining model to address the intricate-style matrix within the outfit sets. Using the Maryland Polyvore and Polyvore Outfit datasets, our approach significantly outperformed state-of-the-art models in text-video retrieval tasks, solidifying its effectiveness in the fashion recommendation domain. This research not only pioneers a new facet of fashion recommendation systems, but also introduces a method that captures the essence of individual style preferences through textual descriptions.
\footnotetext[1]{\vspace{-0.7cm} indicates corresponding author}
\end{abstract}

\section{Introduction}
People generally desire to look stylish, regardless of their level of knowledge or interest in fashion. However, conceiving the perfect outfit without a sense of fashion can be difficult, as a style of clothing consists of a combination of different colors, items, and patterns \cite{lynge2020fashion}. In practice, fashion stylists have been responsible for providing a set of fashion items that match the style of a customer. Thus, outfit recommendation systems in fashion retail have evolved in various ways to meet customers’ desires and create a stylish look.

Typical tasks of outfit recommendation systems include measuring the overall compatibility of multiple fashion items that comprise a complete outfit set (i.e., outfit compatibility prediction) and recommending a fashion item that matches an incomplete outfit set (i.e., complementary item retrieval) \cite{vasileva2018learning, lin2020fashion, cucurull2019context}. Nevertheless, these tasks are limited in that they provide results based on the fashion items pre-selected by customers, which still require some input from the customer's discernment of fashion to finalize the style.

\begin{figure}[t]
    \centering
    \includegraphics[width=0.9\linewidth]{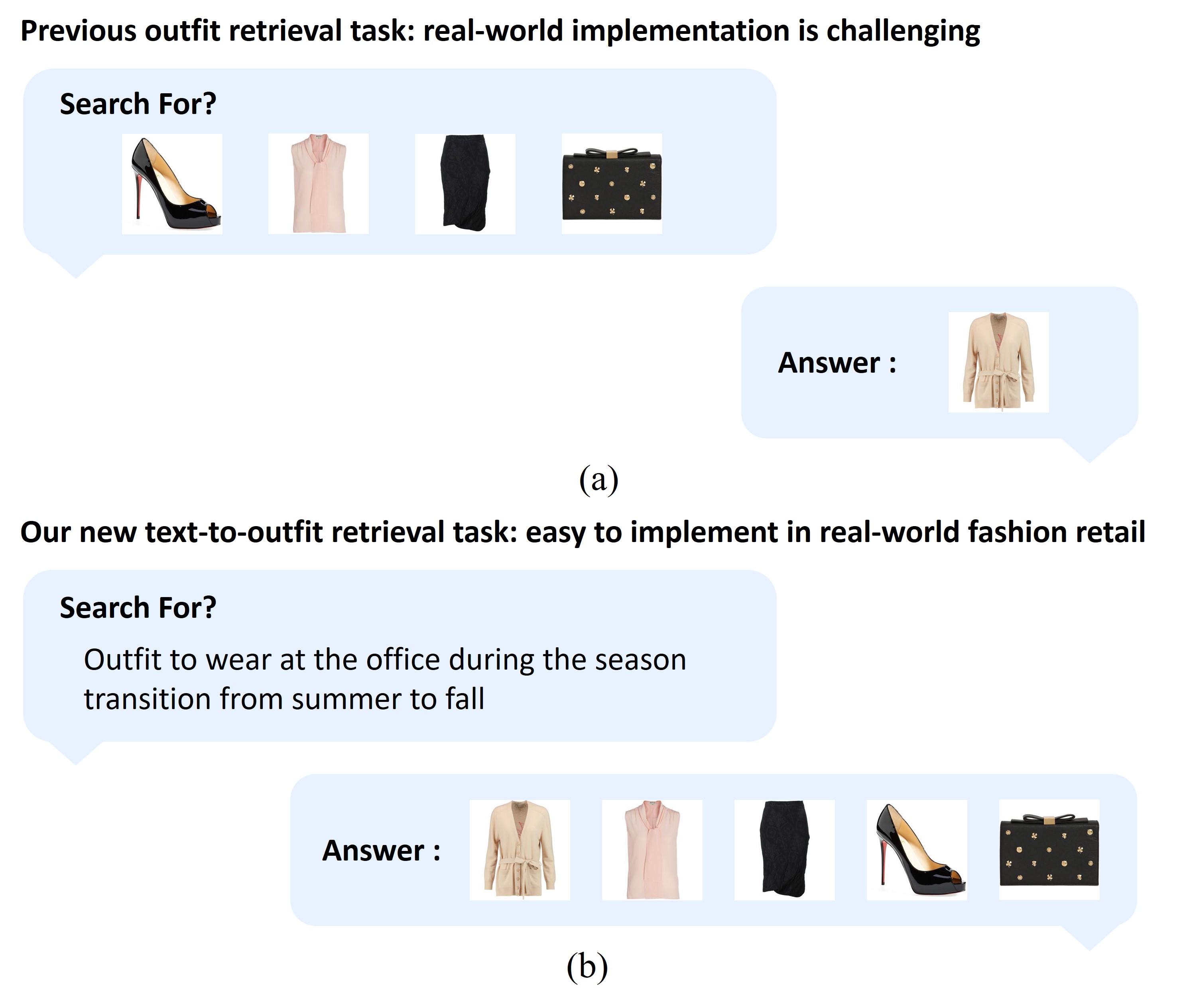}
\caption{Fashion recommendation tasks. (a) Outfit retrieval task completes an outfit by identifying a fashion item that matches a given incomplete set of fashion items. (b) Our new text-to-outfit retrieval task presents a complete outfit set that reflects its textual description given by the customers.}
\label{fig:figure1}
\end{figure}

This study aims to develop a recommendation system that performs the novel task of providing complete styling to customers who are not well-versed with fashion. Considering that consumers input text to search a desired outfit in real-world fashion retail, we propose \textbf{text-to-outfit retrieval task} that recommends an outfit that matches the text entered by the user (\cref{fig:figure1}). Here, the textual description of the outfit is given in sentence form and the recommended outfit is presented as a set of fashion items. As such, the proposed task differs significantly from previous fashion recommendation tasks in that it identifies an appropriate set of fashion items based solely on the customer’s desires, similar to the service provided by a fashion stylist. 

Although this task may be groundbreaking in fashion retail, similar tasks for learning text-image cross representations have progressed in different domains \cite{luo2020univl, zhou2020unified, li2020unicoder, su2019vl}. Representatively, the development of the contrastive language-image pretraining (CLIP) \cite{radford2021learning} model demonstrated impressive results in various vision and language tasks by learning the representation of images and natural language using a vast number of image-caption pairs. In addition to the basic architecture of CLIP, subsequent studies have implemented modules for learning temporal relationships to improve text-video retrieval tasks \cite{fang2021clip2video, luo2022clip4clip, lei2021less}. For example, CLIP4Clip \cite{luo2022clip4clip} took a step further than the CLIP model by implementing a transformer and LSTM to capture the temporal features of a video frame. Accordingly, these aggregated features captured within the sequential data expand the possibilities of text-video retrieval. Further developments in understanding the subtle relationship between text tokens and video frames have been achieved by employing the attention mechanism to model their fine relationships \cite{wang2022disentangled, chen2019weakly, liu2022ts2, jin2022expectation}.

\begin{figure}[t]
    \centering
    \includegraphics[width=0.9\linewidth]{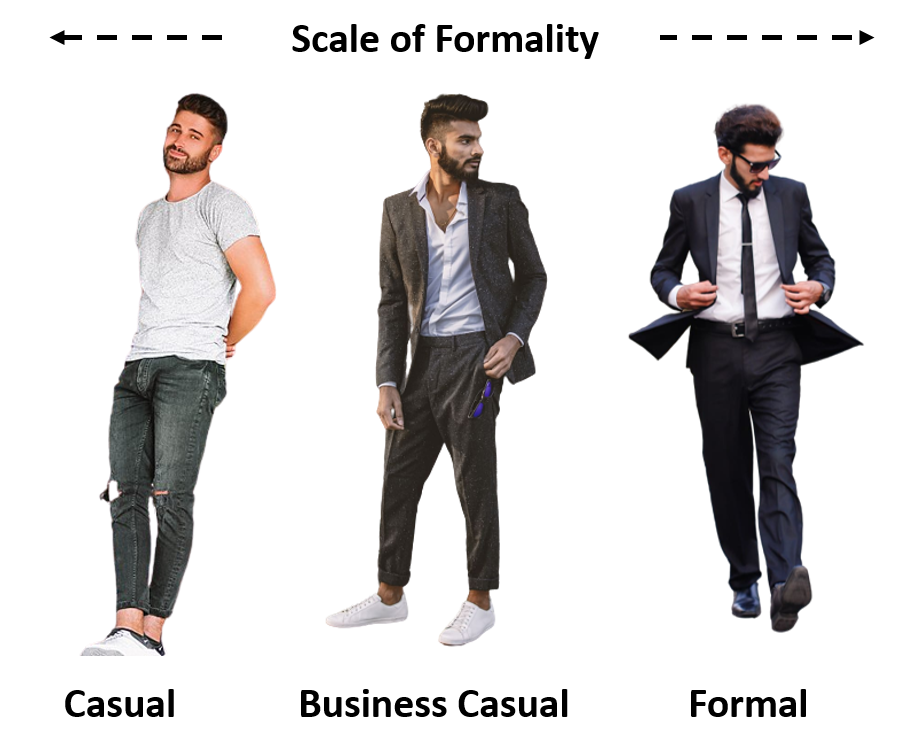}
\caption{Examples of outfits that are similar in colors and patterns but different in styles. The simple combination of a white top and dark pants can vary in style depending on how formal it looks.}
\label{fig:figure2}
\end{figure}
In contrast to text-video models, which primarily focus on modeling temporal features and interactions between adjacent frames of video data, modelling text-to-outfit is significantly different in terms of the unique nature of the outfit data. Given that an outfit is defined as a set of independent fashion items,  no temporal relationship or interconnectivity exists among the items that must be learned by the model. Instead, outfit data have a unique intermediate feature (i.e., style) determined by the combination of various colors, patterns, materials, or prints. While it may be a good idea to take advantage of style-specific characteristics when comprising an outfit, there is no absolute way to determine a well-organized outfit. There are some existing guidelines for what people perceive as stylish, yet it is difficult to perfectly identify appropriate fashion items that reflect the given textual description of a desired outfit. A combination of white shirt and jeans that fits well but differs in style is shown as an example in \cref{fig:figure2}. If the textual description was given to present a casual outfit, the model would have to correctly identify the difference between the styles of the outfit and distinguish it from the formal-looking outfit. That is, the key descriptive words in the text must be identified to match the corresponding item. 

We propose \textbf{CLIP4Outfit}, a novel framework for recommendation systems in the fashion domain that models the correlation between the styles most suitable for a given textual description of an outfit. The inherent relationship between the outfit and text is learned in terms of three semantic levels–item, style, and outfit–where the similarity between the latent information at each level is used for learning. First, a weighted tokenwise interaction (WTI) \cite{wang2022disentangled} module was employed at the item-level to model pairwise correlations between individual fashion items and text tokens. Subsequently, a style extractor module based on a transformer and K-means clustering was devised to encode various styles within an outfit. In particular, to model the outfit similarity at the style-level, each matching text token and fashion outfit were considered as a style category, and the outfit was assumed to consist of multiple style categories. Finally, the outfit-level similarity was computed based on the outfit-level representation derived by aggregating the style-level representations. Consequently, the similarities computed at all three levels were integrated to present the final learning framework.

With this simple but intuitive solution to learn the three semantic levels of outfit, the proposed framework achieved superior results on the text-to-outfit retrieval task using two public datasets: the Maryland Polyvore and the Polyvore Outfit datasets. Specifically, our framework outperformed existing state-of-the-art models in text-video retrieval task, confirming that the proposed learning method is best suited for the fashion recommendation domain. The effectiveness of the proposed learning framework was validated by conducting an ablation study on the proposed style- and outfit-level modules. In addition, a qualitative analysis was performed to analyze how the proposed model employs interactions at each semantic level for recommendations. Finally, we conducted a user study to obtain evaluations that cannot be determined solely by numerical metrics.

The contributions of this study are summarized as follows:
\begin{itemize}
  \item We propose a new fashion recommendation task of text-to-outfit retrieval which identifies a complete set of fashion items based on a given textual description of the customer’s desired outfit.
  \item The proposed model is devised to reflect the interactions between textual description and outfit in three progressive levels–item, style, and outfit. The semantic information of the given text and fashion item images are first encoded at the lowest level (i.e., item) and then constituted into the highest level (i.e., outfit).
  \item The proposed model outperformed the benchmark models used for text-video retrieval tasks. The main components of each semantic level were further validated through ablation study and qualitative analysis.
\end{itemize}

\section{Related Works}
\subsection{Fashion Outfit Compatibility}
Over the years, outfit recommendation systems in the fashion domain have mainly focused on measuring the compatibility of multiple fashion items constituting an outfit \cite{vasileva2018learning, veit2015learning}. A prevalent approach for these algorithms is to adapt the ImageNet architecture to learn the latent feature space for compatibility prediction \cite{lee2017style2vec, veit2015learning, tan2019learning}. In \cite{han2017learning}, Bi-LSTM was used to encode a complete fashion outfit by considering the images of individual fashion items in an outfit as a sequential set of fashion images. Furthermore, subsequent research introduced a representation space with category-specific embeddings to model outfit compatibility using fashion item attributes and categories of fashion items \cite{liu2016deepfashion, vittayakorn2015runway, yang2019interpretable, feng2019interpretable}. In recent studies, learning methods based on graph convolution network \cite{kipf2016semi} have led to commendable outcomes in predicting outfit compatibility. A network of fashion items is constructed by considering the co-occurrence of diverse attributes or categories between fashion items as the weights of the edges in the network \cite{cucurull2019context, zhan20213, cui2019dressing}. However, it is worth noting that predicting outfit compatibility is primarily useful when consumers already have a complete outfit to evaluate. In this sense, the retrieval task of identifying specific fashion items to complete an outfit may be more meaningful than compatibility prediction in terms of a recommendation system that can potentially lead consumers to purchase fashion retail products.

\subsection{Fashion Item Retrieval}
Several recent studies on outfit recommendation systems in the fashion domain have primarily addressed complementary item retrieval. This task aims to recommend suitable fashion items from a large database of fashion items, given an incomplete outfit set \cite{vasileva2018learning, yu2019personalized, ma2020fine}. Such algorithms were first developed based on a general adversarial neural network (GAN), which reconstructs matching items and then identifies similar items within a database \cite{kang2017visually, yu2019personalized}. Moreover, \cite{kang2019complete} constructed an attention map on a scene-to-item level to retrieve complementary items that match the scene. Recently, various frameworks have been proposed to identify complementary items that match a given outfit \cite{lin2020fashion, sarkar2023outfittransformer}. Thus, retrieval tasks in the fashion domain have been focused on searching for fashion items, given a set of fashion-aware images. In contrast, we propose a novel task for an outfit recommendation system in the fashion domain, which aims to search for an outfit set given a description of a particular look that a consumer desires to wear.

\subsection{Text-Video Retrieval}
To effectively execute a text-to-outfit search, our study derives inspiration from methods deployed in text-video retrieval tasks, and structures the methodology in a way that adapts to the fashion domain. In general, recent research on text-video retrieval presents frameworks that efficiently manage text and video inputs through a bi-encoder \cite{liu2019use, radford2021learning, patrick2020support, gabeur2020multi}. Specifically, a common approach among most models is to map text and video data onto a latent space and proceed with the learning process by calculating the similarity between the text and video. A text encoder is typically constructed based on a BERT-like model \cite{liu2019roberta, devlin2018bert}. Conversely, because videos contain a sequence of images, researchers have utilized various multimodal encoders to achieve superior performance \cite{gabeur2020multi, liu2019use, patrick2020support, dzabraev2021mdmmt}. 

Recent studies have demonstrated continual breakthroughs in text-video retrieval using CLIP \cite{radford2021learning}, which learns the semantic similarity between text and video to facilitate the extraction of cross-modal representations. For example, \cite{luo2022clip4clip} trained the CLIP architecture by employing transformers for video encodings to help understand a sequence of photos. Another study \cite{wang2022disentangled} used a token-wise interaction module to enrich the understanding of interactions between video and text data. Furthermore, \cite{liu2022ts2, jin2022expectation} have successfully attempted to accurately model the relationship between text and video by applying techniques such as temporal token shifts and the expectation-maximization algorithm specialized for text-video retrieval. However, attempts to employ CLIP architecture in the fashion domain have been limited to adapting fashion items and descriptions, and have not been used to model outfits. We modelled the text-to-outfit search process by encoding a fashion item sequence into three stages–item, style, and outfit–to achieve a more profound understanding of the relationship between an outfit and its description.

\begin{figure*}[htbp]
    \centering
    \includegraphics[width=\textwidth]{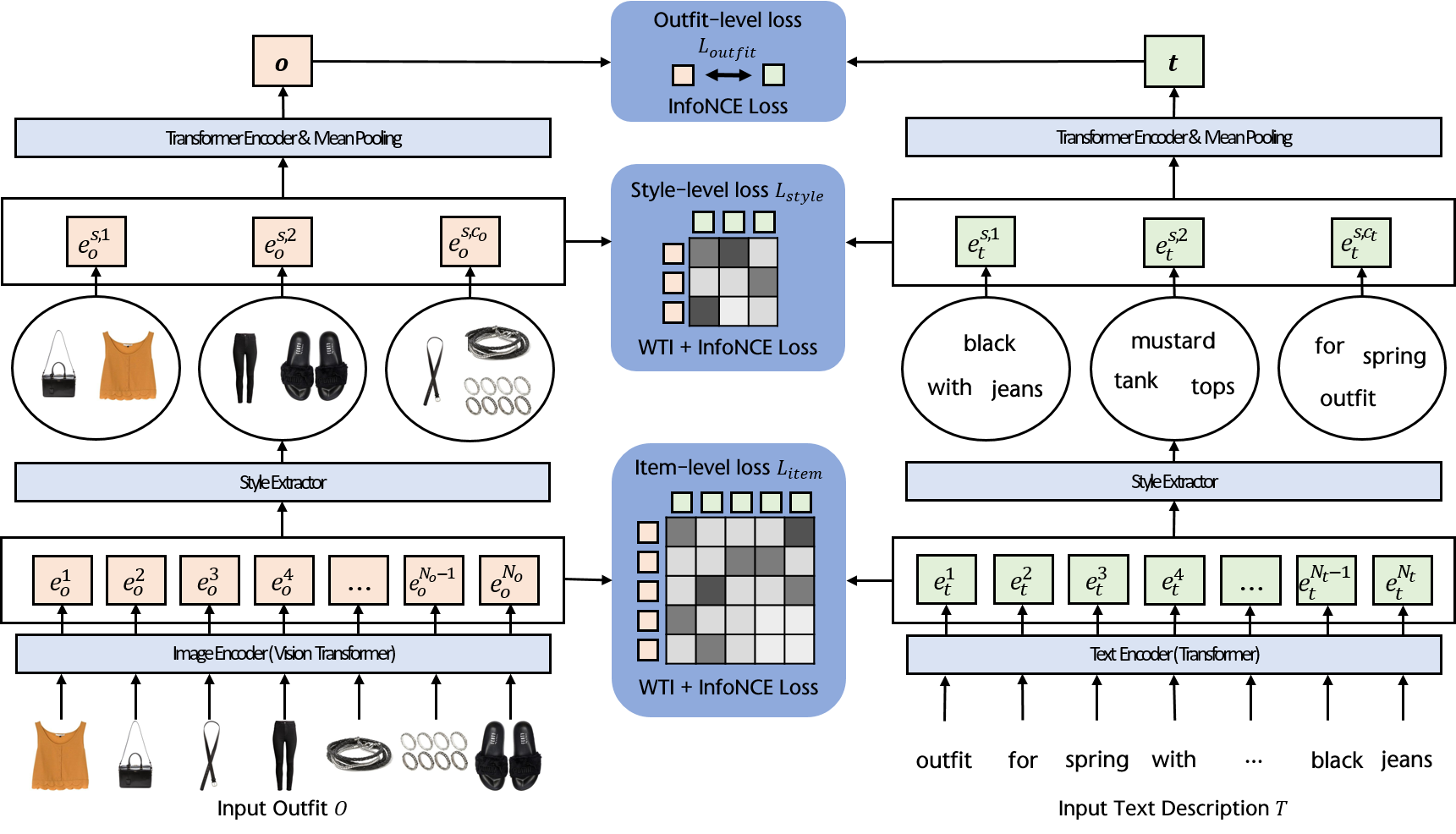}
\caption{Overview of CLIP4Outfit. The proposed model consists of three progressive levels - item, style, and outfit - to learn the semantic information between the text and image of outfits. We use a sentence-long textual description and a set of fashion item images for each outfit as model inputs. First, the item-level images and text tokens are encoded by transformer-based modules to infer the cross-modal similarity between the image and text. Next, style-level similarity is extracted to infer intermediate semantic information as the centroids of text and images through K-means clustering. Finally, outfit-level embedded vectors are computed to capture the global outfit representation.}
\label{fig:figure3}
\end{figure*}

\section{Proposed Model}
The proposed text-to-outfit retrieval framework divides the relationship between the textual description of an outfit and the fashion items constituting the outfit into three levels: item, style, and outfit. The overall architecture of CLIP4Outfit presented in \cref{fig:figure3} was designed to reflect the interactions between the levels. First, at the item-level, correlations between the pairwise features of all fashion items and text tokens were inferred using the model. A WTI module was applied to formalize cross-modal similarity. Next, style-level similarity was computed based on fashion items and text tokens encoded using transformers. The encoded items and tokens were merged using the K-means method and the centroids were subsequently used to extract styles for style-level similarity. Finally, an outfit-level representation was deduced from the extracted styles using transformers to compute the outfit-level similarity. Thus, the three similarity values from each level were appropriately incorporated into InfoNCE loss for model training.
\subsection{Item-level Interaction}\label{section3.1}
Our first goal in text-to-outfit retrieval task is to train the outfit and text encoders through cross-modal representation learning, which involves learning the cross-modal similarity at the item level. First, a text query $T=\{W_{j}\}^{N_{t}}_{1}$ containing $N_t$ tokens and a matching outfit set $O = \{I_{i}\}^{N_{o}}_{1}$ comprising $N_o$ fashion items are provided.  Here, a BERT-like model is used to extract the embedded vector $e_t \in \mathbb{R}^{N_t \times D}$ of text query $T$, whereas a vision transformer is used to extract the embedded vector $e_o \in \mathbb{R}^{N_o \times D}$ of outfit $O$. Subsequently, WTI module is applied to the embedded vectors to learn the cross-modal similarity $S_{O,T}$ defined as follows:
\begin{align}
\begin{split}
S_{O,T} = \cfrac{1}{1+p} \bigg( &\sum\limits_{i=1}^{N_{o}} f_{o}(e_{o})_{i} \max\limits_{j=1}^{N_{t}} \cfrac{(e_{t}^{j})^T e_o^{i}}{\lVert e_{t} \rVert\lVert e_{o} \rVert} + \\
p&\sum\limits_{j=1}^{N_{t}} f_{t}(e_{t})_{j} \max\limits_{i=1}^{N_{o}} \cfrac{(e_{t}^{j})^T e_o^{i}}{\lVert e_{t} \rVert\lVert e_{o} \rVert} \bigg) 
\end{split}
\label{eq1}
\end{align}
where $f_t$ and $f_o$ denote adaptive modules that set the weight of image and text token, respectively, by using $Softmax(linear_t(e_t))$ and $Softmax(linear_o(e_o))$. 
While it is intuitive to train the proposed model based on text-to-outfit similarity for text-to-outfit retrieval task, we also employed outfit-to-text similarity for model training and set appropriate weights between the two similarities. Through our learning approach, we empirically demonstrated that using outfit-to-text similarity in conjunction with text-to-outfit similarity for cross-modal similarity calculations during the training phase enhanced the model performance during the test phase where text-to-outfit similarity is used alone. This finding is discussed in further detail in \cref{section4.4}. Consequently, the contrastive loss $L_{item}$ based on the calculated item-level cross-modal similarity is defined as follows:
\begin{align}
\begin{split}
L_{item} = -\cfrac{1}{B}\left(\sum\limits_{n=1}^{B}\log\cfrac{\exp(S_{O_{n}, T_{n}} / \tau)}{\sum_{m=1}^{B}\exp(S_{O_{n}, T_{m}} / \tau)}\right. \\
\left. + \sum\limits_{m=1}^{B}\log\cfrac{\exp(S_{O_{m}, T_{m}} / \tau)}{\sum_{n=1}^{B}\exp(S_{O_{n}, T_{m}} / \tau)} \right)
\end{split}
\label{eq2}
\end{align}
where $\tau$ and $B$ refer to the temperature parameter and the batch size, respectively.

\subsection{Style-level Interaction}\label{section3.2}
In contrast to traditional text-video retrieval, which optimizes the similarity score at the token- or global-level representation (i.e., item or outfit level in our context), we present a style extractor as an intermediate step between the item- and outfit-levels. Specifically, we developed a module based on transformer and K-means algorithm to extract the myriad styles inherent to individual or various combinations of fashion items.

Initially, the embedded vectors of text $e_t$ and outfit $e_o$ generated at the item level are processed through the transformer encoder $E^{o}_{trans} , E^{t}_{trans} $ as follows:
\begin{align}
\begin{split}
\displaystyle e’_o = E^{o}_{trans}(e_o),  e’_t = E^{t}_{trans}(e_t)
\end{split}
\end{align}

\begin{table*}[h]
\centering
\resizebox{0.65\textwidth}{!}{%
\begin{tabular}{lcccc|cccc}
\hline
\multicolumn{1}{c}{\multirow{2}{*}{Method}} &
  \multicolumn{4}{c|}{Polyvore Outfit} &
  \multicolumn{4}{c}{Maryland Polyvore} \\ \cline{2-9} 
\multicolumn{1}{c}{}                   & R@5  & R@10  & R@30  & R@50  & R@5   & R@10  & R@30  & R@50  \\ \hline
\multicolumn{1}{l|}{SUPPORT-SET}       & 4.13 & 6.71  & 11.29 & 14.76 & 12.44 & 15.72 & 24.31 & 30.38 \\
\multicolumn{1}{l|}{CLIP4Clip} & 6.56 & 9.59  & 17.27 & 21.97 & 15.60 & 20.74 & 29.68 & 35.05 \\
\multicolumn{1}{l|}{Distangle-Net}     & 6.36 & 9.36  & 16.67 & 21.39 & 15.77 & 21.99 & 35.62 & 43.36 \\
\multicolumn{1}{l|}{TS2-Net}           & 6.93 & 9.91  & 17.40 & 22.05 & 16.06 & 22.27 & 36.71 & 44.59 \\
\multicolumn{1}{l|}{EMCL-net}          & 6.74 & 10.10 & 17.53 & 22.63 & 16.42 & 22.49 & 35.85 & 44.87 \\ \hline
\multicolumn{1}{l|}{CLIP4Outfit(Ours)*}              & 7.45 & 10.73 & 18.69 & 24.11 & 18.06 & 24.51 & 39.54 & 47.20 \\
\multicolumn{1}{l|}{CLIP4Outfit(Ours)} &
  \textbf{7.59} &
  \textbf{11.12} &
  \textbf{19.36} &
  \textbf{24.52} &
  \textbf{18.48} &
  \textbf{24.92} &
  \textbf{40.26} &
  \textbf{48.45} \\ \hline
\end{tabular}%
}
\caption{Evaluation of the proposed model against benchmark models on both the Maryland Polyvore and the Polyvore Outfit datasets. * denotes our model trained without pretraining the CLIP module on images and text descriptions of fashion items. }
\label{tab:table1}
\end{table*}

Generally, transformer-based architectures aim to generate classification scores for images or natural languages. Therefore, they are typically used to capture the representations of a single sentence or image. However, our goal in employing transformers within the style extractor was to model the relationship between fashion images and text tokens to define various styles. Here, positional encoding was not required because style is invariant to the order of fashion items and the sequence of text tokens were already considered by CLIP. This is a key difference from transformers in traditional tasks because positional encoding is commonly used to consider a sequential order of images or texts.

Next, the style tokens are extracted from $e’_o$ and $e’_t$ by K-means clustering algorithm. Here, the number of clusters for outfit and text are set as $c_o$ and $c_t$, which is $k_o$\% and $k_t$\% of $N_o$ and $N_t$, respectively. Because the number of tokens to form clusters are minimal compared to typical clustering tasks, randomly initializing the clusters in the K-means algorithm may form clusters that do not reflect the data distribution of token embeddings or produce suboptimal results. In this sense, the initial cluster centroids of the algorithm were set by greedy selection based on the distance between the data points, where the detailed mechanism and its effect are demonstrated in \cref{alg:cap} and \cref{section4.4}. Accordingly, the two sets of cluster centroids (i.e., $e^s_o$ and $e^s_t$) extracted using the normalized embedded vector are treated as token vectors at the style-level that have a higher semantic level than the item-level. By computing the cross-modal similarity between $e^s_o$ and $e^s_t$, the contrastive loss $L_{style}$ was applied in the same manner as at the item-level.

\begin{algorithm}[t]
\caption{Style extractor algorithm}\label{alg:cap}
\begin{algorithmic}

\Require for $x \in \{t, o\}$ , item-level embedded vectors $e_{x}$, transformer encoder $E^{x}_{trans}$ , k-means ratio $k_{x}$
\Ensure centroid set $e^s_{x}$
\Algorithm{}
\State $c_{x} \gets \max( int (k_{x}*len(e_{x}), 1)$
\State $e_{x} \gets E^{x}_{trans}(e_{x})$
\State \text{\ttfamily /* Select the first centroid from $e_{x}$ */}
\State $e^s_{x} \gets \argmax\limits_{e_{x}^{i} \in e_{x}}\sum\limits_{i \neq j} \lVert e_{x}^{i} - e^{j}_{x} \rVert_{2} $
\State
\State \text{\ttfamily /* Apply greedy selection  */}
\For{$j \in  [2,3 ,... c] $}
    \State $c_{j} \gets \argmax\limits_{e_{x}^{i} \in e_{x}}\min\limits_{C \in e^{s}_{x}}\lVert e_{x}^{i} - C \rVert_{2}$
    \State $e_{x}^{s} \gets e_{x}^{s} \cup \{c_j\}$
\EndFor
\State $e_{x}^{s} \gets K$-$Means(e_{x}, e_{x}^{s})$
\State \Return $e^{s}_{x}$
\end{algorithmic}
\end{algorithm}

\subsection{Outfit-level Interaction}

Given that the semantic information of an outfit is presumed to be a combination of the relationships between different styles, we define the outfit-level embedded vectors $\textbf{o}, \textbf{t}$ from the extracted style tokens $e^s_o$ and $e^s_t$ as follows: 
\begin{align}
    \begin{split}
        \displaystyle \textbf{o} = &MeanPool(F^{o}_{trans}(e^s_o))\\
        \displaystyle \textbf{t} = &MeanPool(F^{t}_{trans}(e^s_t)).
    \end{split}
\end{align}

Here, $e^s_o$ and $e^s_t$ are processed through the transformer encoder $F^{o}_{trans}$ and $F^{t}_{trans}$, where the relationships between different styles are modeled by mean pooling and integrated into $\textbf{o}, \textbf{t}$. Inspired by the CLIP model, the similarity function $S^{o}_{\textbf{o}, \textbf{t}}$ between outfits $\textbf{o}$ and $\textbf{t}$ is devised using the cosine function because the outfit-level embedded vector infers a global representation. The contrastive loss $L_{outfit}$ is defined by calculating InfoNCE with the similarity $S^{o}_{\textbf{o}, \textbf{t}}$ , making it analogous to $L_{item}$ and $L_{style}$. Thus, the training objective is stated as follows: 
\begin{align}
    \begin{split}
\displaystyle L = & L_{item} + \alpha L_{style}+ \beta L_{outfit}
    \end{split}
\end{align}
where $\alpha$ and $\beta$ are the weighting parameters.

In addition, we pretrained the CLIP model on text-image tasks to use the parameters of the pretrained model for initializing the text and outfit encoders used in this task. This facilitates the image encoder in capturing fashion-specific features and extracting a better outfit representation. 

\section{Experiments}
\subsection{Dataset}
Two outfit datasets from Polyvore were used to train and evaluate the proposed model. For every outfit in the datasets, the image and textual description of each fashion item included in the outfit set, and the overall textual description of the outfit were provided.

\begin{itemize}
  \item Maryland Polyvore \cite{han2017learning}: This is a crowdsourced dataset created by users of the Polyvore website. The collection of images and textual descriptions of the fashion items in the outfits are uploaded by the users. This dataset consists of 164,379 fashion items, which are split into 17,316, 1,497, and 3,076 outfit sets for training, validation, and testing, respectively.
  \item Polyvore Outfit \cite{vasileva2018learning}: This is a larger dataset from the Polyvore website. It comprises 365,054 fashion items, which are divided into 53,306, 5,000, and 10,000 outfit sets for training, validation, and testing, respectively.
\end{itemize}
\subsection{Implementation Details}
To learn the image and textual descriptions of the outfits, we employ the standard biencoder of the CLIP model. Specifically, ViT-B/32 \cite{dosovitskiy2020image} was used as the image encoder and a transformer-based architecture was used as the text encoder. These encoders represent the image and text data in 512 dimensions. In addition, only one layer of the transformer was stacked for each encoder used to model the relationship between the style and outfit level.

Given that the number of clusters in K-means algorithm is based on the number of items included in an outfit, which ranges from 4 to 19, we use a ratio of 1/3 for k-means clustering to capture outfit styles and a ratio of 1/6 for capturing text styles. Moreover, Adam optimizer with initial learning rate of $5\times10^{-7}$ was used to train the model with batch size set as 50 and all learning rates were trained according to the cosine annealing schedule for 20 epochs. For the loss function, we set the temperature parameter $\tau$ as 100, style level loss weight $\alpha$ as 0.3, outfit level loss weight $\beta$ as 0.3, and task weight $p$ as 0.2.

\subsection{Comparison with State-of-the-art}
We compared our text-to-outfit retrieval model with state-of-the-art methodologies in text-video retrieval. The text-to-outfit retrieval performance was evaluated on the entire test outfit set using recall@top-k metric for k values of 5, 10, 30, and 50 (i.e., R@5, R@10, R@30, R@50). The benchmark models used for comparison were SUPPORT-SET\cite{patrick2020support}, CLIP4Clip\cite{luo2022clip4clip}, TS2-Net\cite{liu2022ts2}, Distangle-Net\cite{wang2022disentangled}, and EMCL-Net\cite{jin2022expectation}. In addition, for precise comparison, we reported both cases of our model where the CLIP module was and was not pretrained on the image and text descriptions of fashion items.

\cref{tab:table1} demonstrates the superior performance of our model across all metrics compared to benchmark models and even when the model is not pretrained on CLIP. CLIP-based benchmark models outperformed Support-Set, inferring that employing CLIP for multi-modal learning contributed significantly to the model performance.  Furthermore, the proposed model outperformed TS2-Net and EMCL-net, which were specialized for text-video tasks by applying the interaction of tokens.  Notably, our model achieved 24.92\% in the R@10 index on the Maryland Polyvore set, which is a 10.8\% improvement over the next-best-performing benchmark model.

\begin{table*}[t]
    \centering
    \begin{minipage}{.5\textwidth}
        \centering
        \resizebox{0.95\linewidth}{!}{%
            \begin{tabular}{@{}c|cccc|cccc@{}}
            \toprule
            \multirow{2}{*}{Method} & \multicolumn{4}{c|}{Polyvore Outfit} & \multicolumn{4}{c}{Maryland Polyvore} \\ \cmidrule(l){2-9} 
                                    & R@5     & R@10    & R@30    & R@50    & R@5     & R@10    & R@30    & R@50    \\ \midrule
            I+S+O                   & 7.59    & 11.12   & 19.36   & 24.52   & 18.48   & 24.92   & 40.26   & 48.45   \\
            I+S                     & 6.96    & 10.20   & 18.07   & 23.37   & 17.95   & 24.24   & 38.86   & 46.59   \\
            I                       & 6.49    & 9.51    & 17.37   & 22.24   & 16.39   & 23.41   & 36.70   & 44.75   \\ \bottomrule
            \end{tabular}%
            }
        \caption{Results of ablation study. I, S, and O denote item-, style-, and outfit-level modules, respectively.}
        \label{tab:table2}
    \end{minipage}%
    \begin{minipage}{.5\textwidth}
        \centering
        \resizebox{0.95\linewidth}{!}{
                \begin{tabular}{@{}ccccccccc@{}}
                \toprule
                \multirow{2}{*}{Method}   & \multicolumn{4}{c}{Polyvore Outfit}              & \multicolumn{4}{c}{Maryland Polyvore} \\ \cmidrule(l){2-9} 
                                          & R@5  & R@10  & R@30  & \multicolumn{1}{c|}{R@50}  & R@5     & R@10    & R@30    & R@50    \\ \midrule
                \multicolumn{1}{c|}{No greedy}           & 7.36 & 10.89 & 19.08 & \multicolumn{1}{c|}{24.05} & 18.33 & 24.62 & 39.95 & 47.39 \\
                \multicolumn{1}{c|}{\centering No $o$-$t$ ($p=0$)} & 7.41 & 10.97 & 19.23 & \multicolumn{1}{c|}{23.96} & 18.30 & 24.74 & 39.33 & 47.75 \\
                \multicolumn{1}{c|}{1:1 $o$-$t$ ($p=1$)}      & 6.88 & 10.61  & 18.32 & \multicolumn{1}{c|}{23.71} & 17.83 & 24.71 & 38.44 & 46.73 \\
                \multicolumn{1}{c|}{Ours} & 7.59 & 11.12 & 19.36 & \multicolumn{1}{c|}{24.52} & 18.48   & 24.92   & 40.26   & 48.45   \\ \bottomrule
                \end{tabular}%
                }
        \caption{Results of parameter sensitivity. No greedy refers to random initialization of K-means and $o$-$t$ refers to outfit-to-text similarity.}
        \label{tab:table3}

    \end{minipage}
\end{table*}


This significant improvement compared to text-video models on both fashion outfit datasets can be attributed to the unique factors of outfit data. In case of video data, which is a sequence of images over time, learning its representation requires the consideration of the temporal relationship between images (i.e., positional encoding). In contrast, the text-to-outfit retrieval task does not need to consider temporal relationships because a set of fashion items in an outfit do not involve sequential information. Furthermore, given that sequential images in video data do not usually change dramatically, intermediate modules (i.e., style extractors) based on K-means clustering may not be effective, because these images are likely to be distributed tightly around the centroids. Conversely, an outfit set consisting of various fashion items can be highly favorable for the novel approach of modeling semantic information within the outfit set step-by-step at the style- and outfit-levels.

\subsection{Ablation Study and Parameter Sensitivity}\label{section4.4}
We conducted an ablation study to validate the main components within each level of text-to-outfit retrieval in the proposed model. \cref{tab:table2} demonstrates that all components at the item-, style-, and outfit-levels contributed equally to the results of the text-to-outfit retrieval. Specifically, utilizing a similarity matrix with style tokens extracted from fashion items resulted in a significantly improved performance in R@10 compared to the baseline model by a maximum of 7.2\%. Similarly, outfit-level aggregation led to a performance improvement of 9\%. 
\begin{figure}[b]
    \centering
    \begin{minipage}{0.48\linewidth}
        \includegraphics[width=\linewidth]{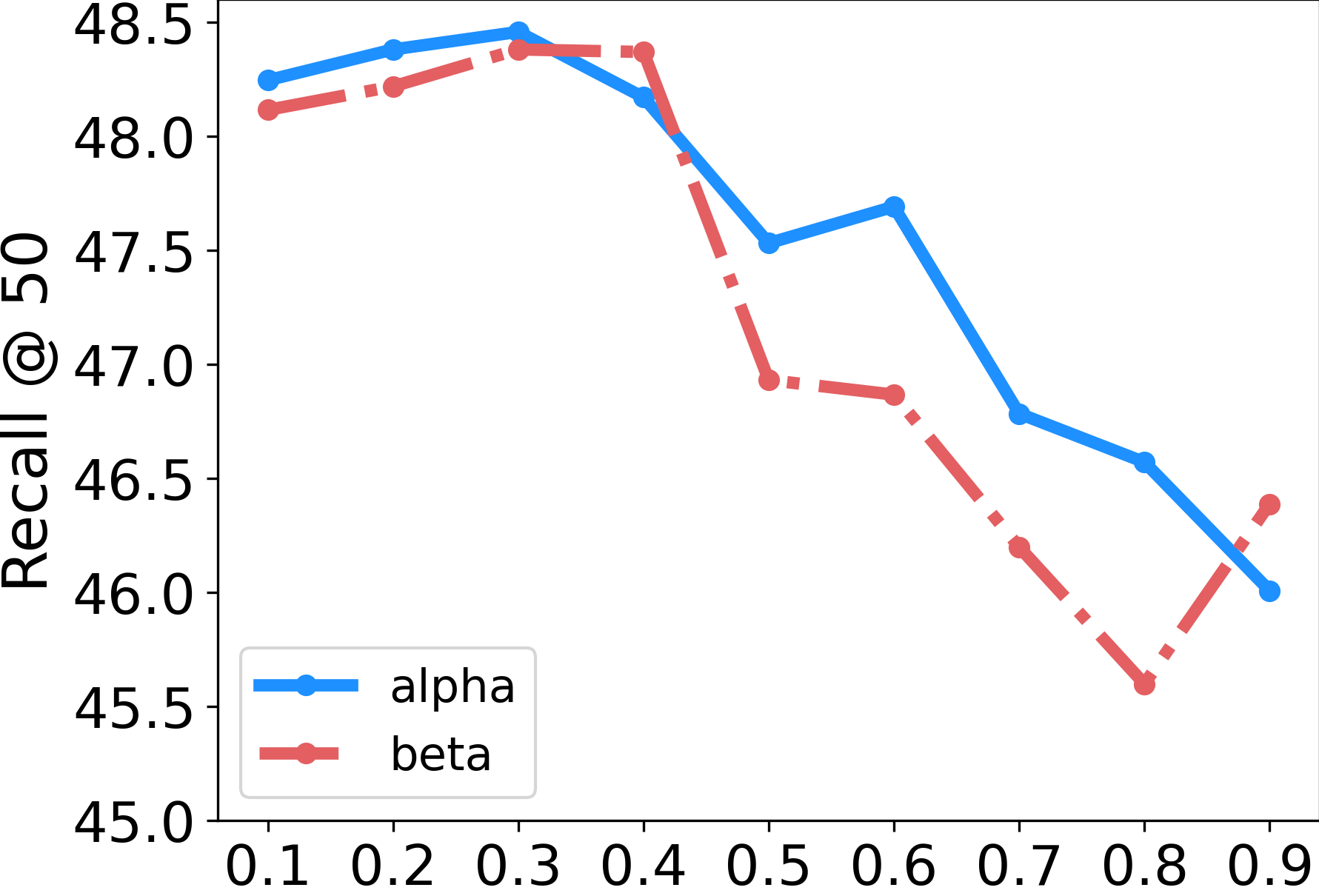}
        \caption{Evaluation of loss parameters $\alpha$ and $\beta$}
        \label{fig:figure4}
    \end{minipage}
    \hfill
    \begin{minipage}{0.48\linewidth}
        \includegraphics[width=\linewidth]{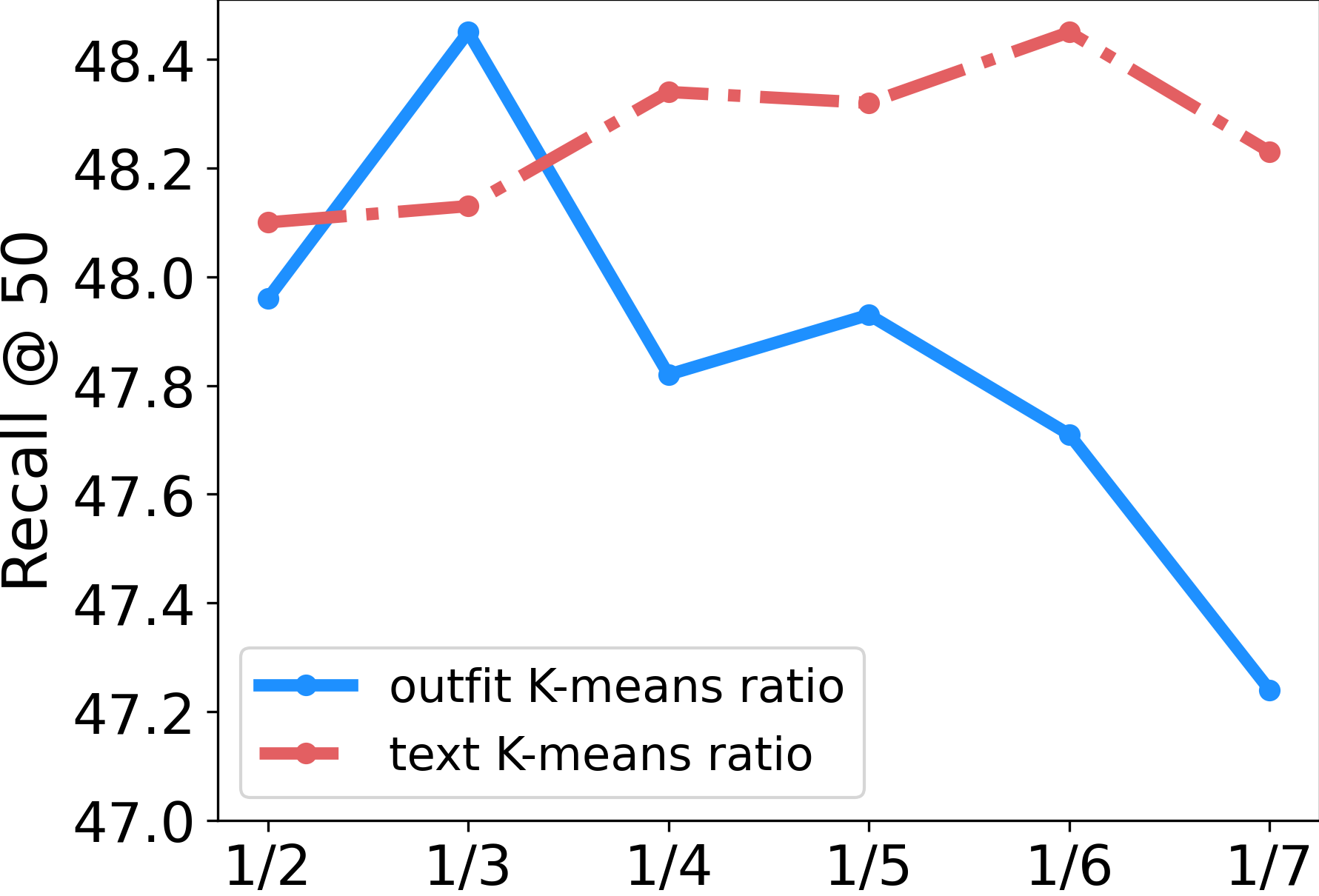}
        \caption{Evaluation of centroid ratios for outfit and text}
        \label{fig:figure5}
    \end{minipage}
\end{figure}

Given that various parameters are involved within the model, we evaluated parameter sensitivity by gradually adjusting the hyperparameters to identify the optimal design for each parameter. First, the weight parameters of the loss functions, $\alpha$ and $\beta$, were evaluated by altering the values on a scale of 0.1 to 0.9. The optimal result of $\alpha$ and $\beta$ are both reported as 0.3 in \cref{fig:figure4}. Next, the ratio for determining the number of k-means centroids in the style extractor was assessed from 1/2 to 1/7 in \cref{fig:figure5}, where extracting style with too few or too many clusters was found to be detrimental to the learning results. Furthermore, we assessed the efficacy of our greedy centroid initialization method proposed in \cref{section3.2} by contrasting it with the K-means algorithm using random initialization. \cref{tab:table3} reports our method of using greedy selection for initialization as 2.3\% increase in R@50. Finally, we assessed the weight $p$ for outfit-to-text similarity by adjusting it on a scale from 0 to 1. The performance gradually increased from not using outfit-to-text similarity (i.e., $p$=0) and then decreased, with the optimal weight found at 0.2. In \cref{tab:table3}, we have listed the results for the standard retrieval case with $p$=1 and for the text-to-outfit retrieval scenario where it seems intuitive to use $p$=0.

\begin{figure}[b]
    \centering
    \includegraphics[width=0.95\linewidth]{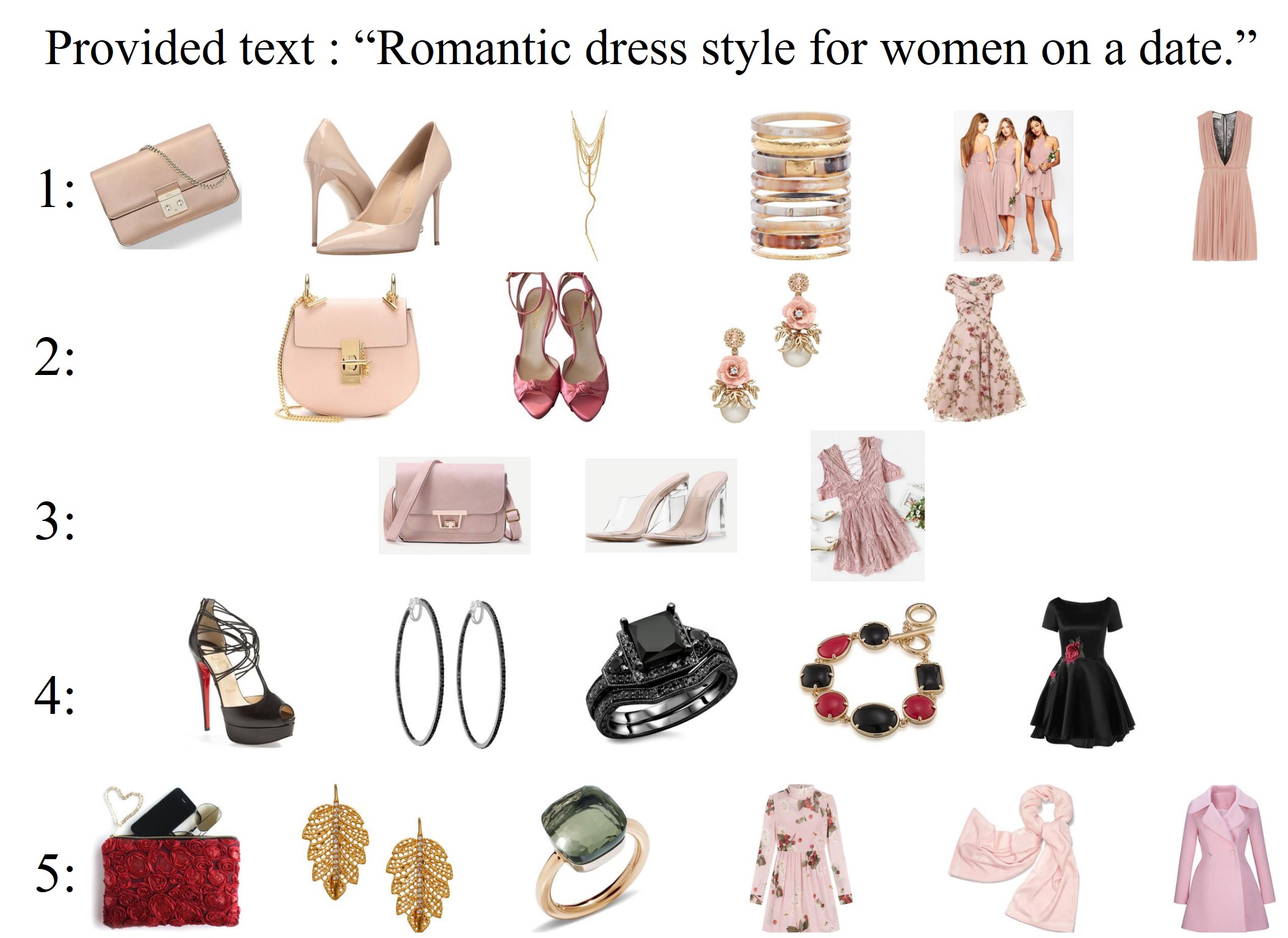}
\caption{Sample question in user study. Five candidate outfit sets that match the textual description are presented to the annotators.}
\label{fig:figure6}
\end{figure}

\begin{figure*}

    \centering
    \includegraphics[width=0.78\linewidth]{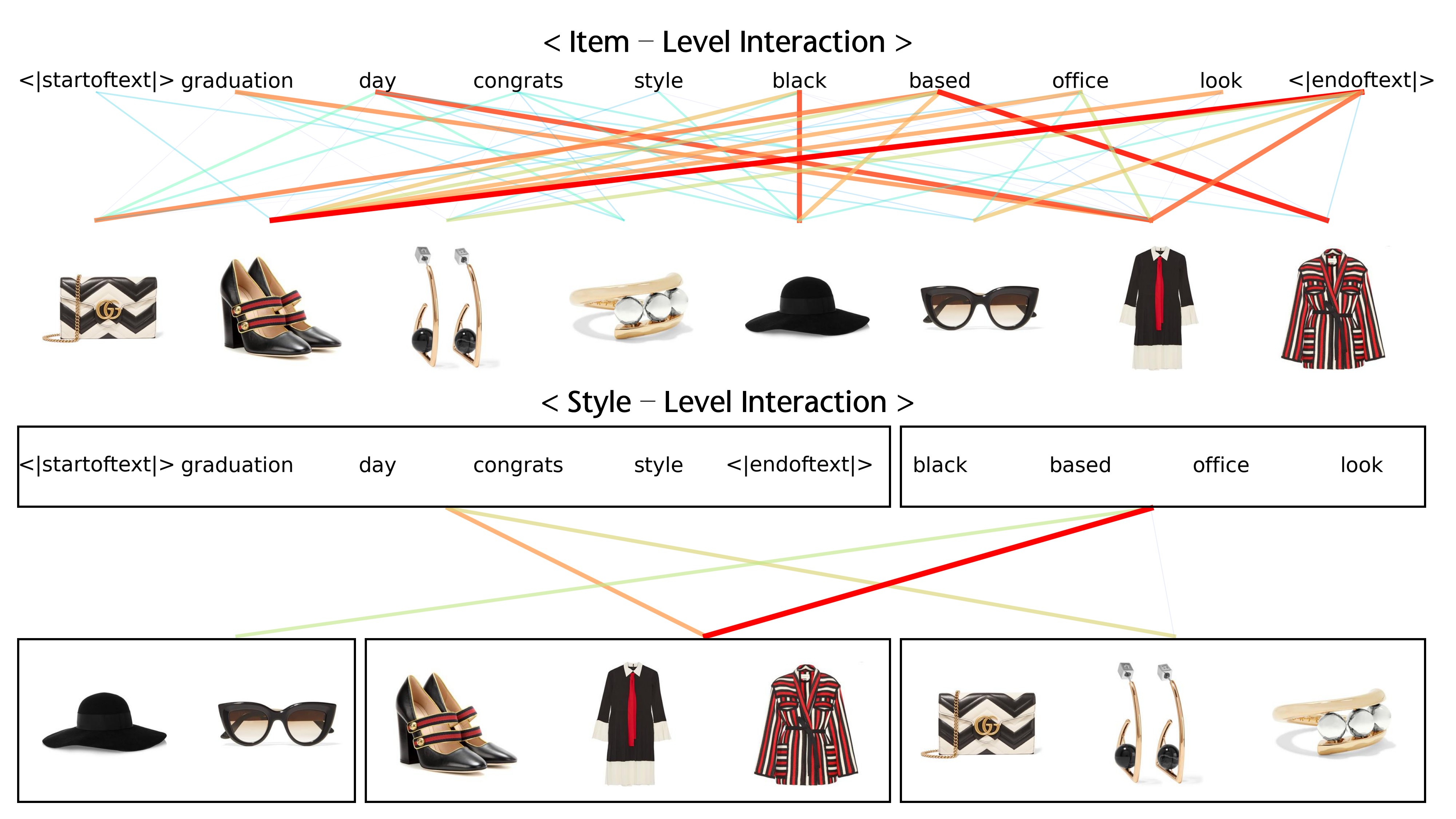}
\caption{Visualized result of pairwise token interactions at item- and style-levels for qualitative analysis.}
\label{fig:figure7}
\end{figure*} 

\subsection{User Study}
While the proposed model demonstrates remarkable recall scores in both datasets, it is important to note that fashion style is inherently subjective. In addition, as the text-to-outfit retrieval task is newly proposed in this study, we conducted a user study with potential consumers to intuitively evaluate the recommendation results, which is limited by recall score alone. Accordingly, user study was performed in two ways using Amazon Mechanical Turk to assess the recommended outcomes. For quality control measures in both cases, annotators who had accomplished a minimum of 1,000 tasks and had exhibited $>$98\% acceptance rate in their prior tasks were eligible to participate. 

First, the annotators were presented with five candidate outfit sets that matched the given textual description for each question and were asked to select whether they were satisfied with the recommended outfits and to rate their satisfaction on a scale of 1 to 10. A total of 5,340 questions were asked, where the sets of text-to-outfit presented to the annotators are shown in the Appendix and \cref{fig:figure6}. Here, the given text queries were generated with ChatGPT to reflect various range of styles. A notable result was reported as a significant majority of 91.17\% of the annotators expressed satisfaction with the suggested outfits. In addition, the average satisfaction rate remarkably stood at 7.619, underlining the positive response to the recommendations.

The second user study was executed based on the text queries within the test set to perform an A/B test between the outcome of our model and the ground truth. A total of 5,180 questions were asked and 51.93\% of the annotators selected the items suggested by our model as more favorable than the ground truth. By assessing customer preference through two ways that may reflect real-world scenarios within the fashion retail domain,  the demonstrated strong inclination towards our model underscores its potential for effective implementation.

\subsection{Qualitative Analysis}
To underscore the significance of our findings, we present a graphical summary of how interactions at the item- and style-levels contributed to the predicted results. At the item-level similarity, interaction was significantly apparent between the text describing the individual item and the corresponding images of the items. For example, strong interaction was observed between the word \emph{black} and the image of a black hat in \cref{fig:figure7}. Similarly, in case of style-level similarity, an active interaction was noted between the clusters of fashion items and the clusters of text tokens that described the common features of fashion item clusters. The visual sample in \cref{fig:figure7} exemplifies this by showing how a group of text tokens like \emph{graduation, day, congrats, style} align semantically with images of formal attire. Remarkably, certain text tokens that showed limited item-level interactions, such as congrats and style, gained prominence when grouped within the style-level context. In essence, our adeptly captures the unique similarity between outfit and text tokens, operating seamlessly across different levels of semantic abstraction. Thus, visualizing the interactions at each level can be used as both an explanation of the principle behind the proposed model and a justification for successfully devising our modules to observe the interactions within the three levels.

\section{Conclusion}
We present a new text-to-outfit retrieval task for a fashion outfit recommendation system and propose a novel modeling method for learning cross-modal representations between text and fashion outfits. Specifically, we defined a fashion outfit as a combination of different styles, whereas style is a combination of different items. Thus, the style extractor module was devised to derive certain styles from a set of fashion items. Although it is difficult to model all precise inherent relationships between text and outfit, we implemented the weighted token interaction module and clustering methods to learn the token-by-token interactions at each semantic level, achieving state-of-the-art performance on various datasets in the fashion domain. In addition, our model is also prominent in that it provides a visualization of how tokens interact at each semantic level. Given that the architecture of the proposed model is based on transformers, we expect the performance of the model to improve with the increased availability of text-to-outfit pairs of good quality.

{\small
\bibliographystyle{ieee_fullname}
\bibliography{egbib}
}

\end{document}